\documentclass{article}

\usepackage[nonatbib,final]{n}

\usepackage[utf8]{inputenc} 
\usepackage[T1]{fontenc}    
\usepackage[bookmarks=false,colorlinks=true,linkcolor=black,citecolor=black,filecolor=black,urlcolor=black]{hyperref}
\usepackage{url}            
\usepackage{booktabs}       
\usepackage{amsfonts}       
\usepackage{nicefrac}       
\usepackage{microtype}      

\usepackage{graphicx}
\usepackage{multirow}

\usepackage{tabulary}

\usepackage{amsmath,amsthm}

\usepackage[british,UKenglish,USenglish,english,american]{babel}

\title{S-OHEM: Stratified Online Hard Example Mining for Object Detection}

\author{
  Minne Li \\
  School of Computer, NUDT \\
  Changsha, China 410073 \\
  \texttt{liminne15@nudt.edu.cn} \\
  \And
  Zhaoning Zhang \\
  School of Computer, NUDT \\
  Changsha, China 410073 \\
  \texttt{zzningxp@gmail.com} \\
  \thanks{This work was done by Minne Li as a student, and Zhaoning Zhang is the corresponding author.}
  \And
  Hao Yu \\
  School of Computer, NUDT \\
  Changsha, China 410073 \\
  \texttt{hyunudt@gmail.com} \\
  \AND
  Xinyuan Chen \\
  School of Computer, NUDT \\
  Changsha, China 410073 \\
  \texttt{cxypdl@gmail.com} \\
  \And
  Dongsheng Li \\
  School of Computer, NUDT \\
  Changsha, China 410073 \\
  \texttt{dsli@nudt.edu.cn} \\
}

\begin{document}

\maketitle
\begin{abstract}
One of the major challenges in object detection is to propose detectors with highly accurate localization of objects. The online sampling of high-loss region proposals (hard examples) uses the multitask loss with equal weight settings across all loss types (e.g, classification and localization, rigid and non-rigid categories) and ignores the influence of different loss distributions throughout the training process, which we find essential to the training efficacy. In this paper, we present the {\em Stratified Online Hard Example Mining (S-OHEM)} algorithm for training higher efficiency and accuracy detectors. S-OHEM exploits OHEM with stratified sampling, a widely-adopted sampling technique, to choose the training examples according to this influence during hard example mining, and thus enhance the performance of object detectors. We show through systematic experiments that S-OHEM yields an average precision (AP) improvement of 0.5\% on {\em rigid categories} of PASCAL VOC 2007 for both the IoU threshold of 0.6 and 0.7. For KITTI 2012, both results of the same metric are 1.6\%. Regarding the mean average precision (mAP), a relative increase of 0.3\% and 0.5\% (1\% and 0.5\%) is observed for VOC07 (KITTI12) using the same set of IoU threshold. Also, S-OHEM is easy to integrate with existing region-based detectors and is capable of acting with post-recognition level regressors.
\end{abstract}
\section{Introduction}
One of the major and fundamental challenges in object detection is to increase localization accuracy, which indicates the detector's ability to predict correct regions of target objects. The metric is typically measured by the bounding box overlap, i.e., the intersection over union (IoU) of the ground truth and predicted bounding boxes. While previous challenges (e.g, PASCAL VOC ~\cite{everingham2010pascal} and KITTI~\cite{Geiger2012CVPR}) normally requires an IoU threshold of 0.5 to be considered a correct detection, real-world applications usually call for a higher accuracy (e.g, IoU $\geq$ 0.7). For example, the vehicle and pedestrian detection in autonomous driving need an accurate measurement of distance through real-time road traffic captures.

Recent literature has focused on the modification of region-based detection models at the post-recognition level to boost the localization accuracy~\cite{felzenszwalb2010object,gidaris2016locnet,gidaris2015object}. However, limited work has addressed the problem from a data perspective. Data is important. The rapid advancement in the data collection, storage, and processing technology has made machine learning, especially deep learning, much easier by lightening the burden of generalizing well to unseen data with a limited number of training data~\cite{Goodfellow-et-al-2016}.

However, the challenge of learning from imbalanced data~\cite{he2009learning} still exists. Within the "Recognition Using Regions" paradigm~\cite{gu2009recognition}, the training set of object detection is divided into two distinct groups of annotated objects and background regions, and the number of examples in these groups experience a huge imbalance. Online Hard Example Mining (OHEM)~\cite{shrivastava2016training} is proposed to overcome the data imbalance by integrating bootstrapping technique~\cite{Sung:1996:LES:929901} with region-based detectors, and can be effortlessly implemented on most of the region-based detectors.

\begin{figure}
\begin{center}
\includegraphics[height=5cm]{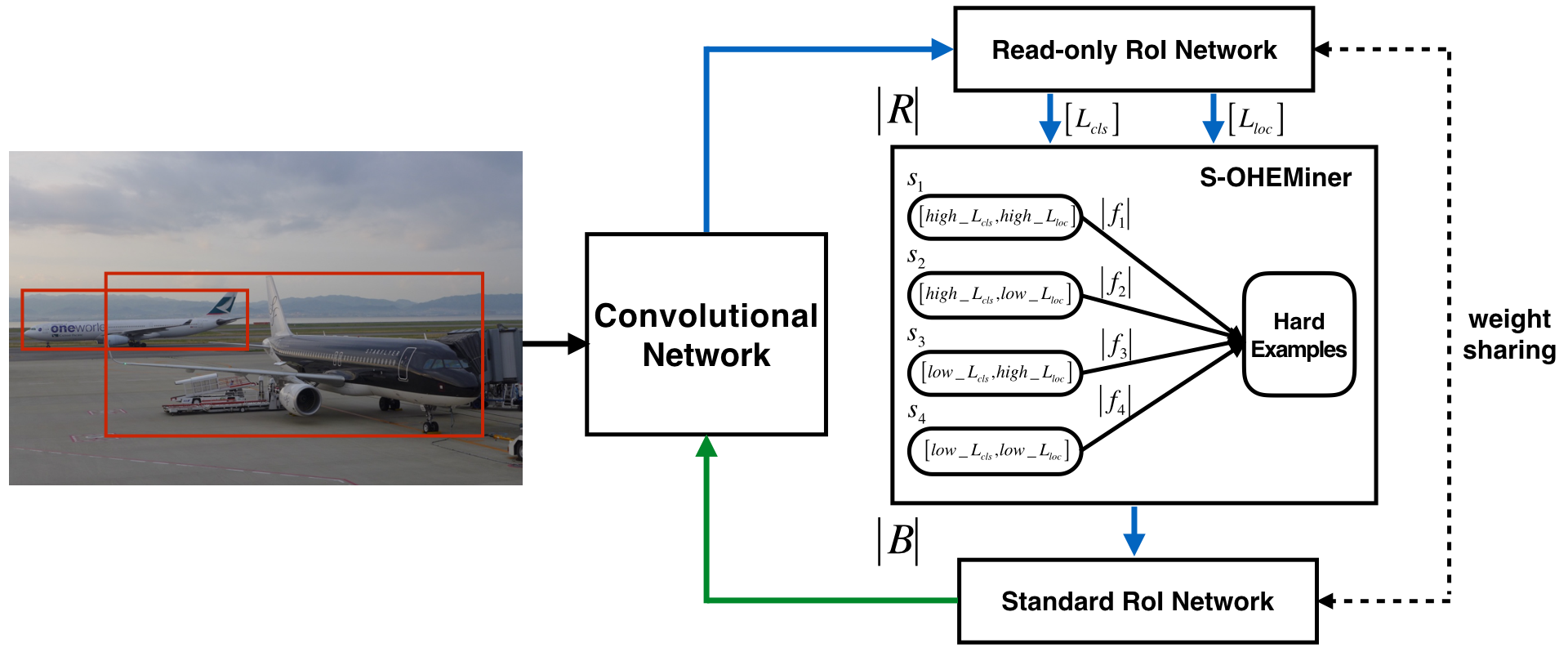}
\end{center}
   \caption{{\bf Architecture of the Stratified Online Hard Example Mining algorithm (S-OHEM).}  
   We use the parameter denotation from~\cite{shrivastava2016training}. In each mini-batch iteration, $N$ is the number of images sampled from the dataset, $R$ is the number of forward-propagated RoIs, and $B$ is the number of subsampled RoIs to be fed into backpropagation. We denote classification loss by $L_{cls}$ and localization loss by $L_{loc}$. S-OHEMiner conducts stratified sampling over $R$ region proposals according to the sampling distribution at current training stage and produces $B$ RoIs to be fed into backpropagation. We maintain a read-only RoI network and a standard RoI network with sharing weights for efficient memory allocation, derived from \cite{shrivastava2016training}. The blue solid stream indicates the process of forward-propagation and the green dashed stream shows the backpropagation process. More details are described in Sect. 3.3.}
\label{fig:S-OHEM}
\end{figure}

In this paper, we propose S-OHEM, the {\em Stratified Online Hard Example Mining} algorithm for training region-based deep convolutional network detectors to enhance localization accuracy, as shown in Fig. 1. The intuition of our method is that feeding hard examples to the backpropagation process could overcome the dilemma of unbalanced data, resulting in a more efficient and effective training process~\cite{shrivastava2016training}. In the field of object detection, the hard example is defined as region proposal with higher training loss. Thus, previous hard example mining method (e.g, OHEM) is conducted by sampling region proposals according to a distribution that favors high loss instances. However, the training loss defined in previous work is the multitask loss with equal weight settings across all loss types (e.g, classification, localization, mask~\cite{he2017mask}, or rigid categories and non-rigid categories). This approach ignores the influence of different loss types throughout the training process, which we found essential to the training efficacy (e.g, localization loss is more important during the latter part of the training period). Therefore, maintaining a sampling distribution according to this influence during hard example mining should enhance the performance of object detectors.

S-OHEM exploits {\em stratified sampling}, a sampling method involving the division of a population into distinct groups known as {\em strata}~\cite{li2016dss} (homogeneous subgroups, in which the inner items are similar to each other). During each mini-batch iteration, S-OHEM firstly assigns candidate examples (in the form of Region of Interests, RoIs) to different strata by the ratio between classification and localization loss. Then the RoIs are subsampled according to a dynamic distribution and fed into the backpropagation process. With an increasing focus on the localization loss, S-OHEM can predict more accurate bounding boxes and therefore enhance the localization accuracy. We apply S-OHEM to the standard Fast R-CNN~\cite{girshick2015fast} and Faster R-CNN~\cite{ren2015faster} detection method and evaluate it on PASCAL VOC 2007 and KITTI datasets. Our systematic experimental analysis reports that S-OHEM yields some AP improvements of 0.5\% on {\em rigid categories} of PASCAL VOC 2007 for both the IoU thresholds of 0.6 and 0.7. For KITTI 2012, both results of the same metric are 1.6\%. Regarding the mAP, a relative increase of 0.3\% and 0.5\% (1\% and 0.5\%) is observed for VOC07 (KITTI12) with the same set of IoU threshold.

The remainder of this paper is structured as follows. In Sect. 2, we compare our work with related research with a focus on the improvement of localization accuracy and the use of data in object detection. In Sect. 3, we describe the design of the algorithm. In Sect. 4, we show the experimental results, and in Sect. 5, we conclude this work.

\section{Related Work}
Object detection has significantly benefited from the advancement of image classification task. The remarkable feature extraction ability of Deep Convolutional Networks~\cite{krizhevsky2012imagenet,szegedy2015going,SimonyanZ14a,he2016deep,szegedy2016rethinking} has equipped us with abundant information for the classification of region proposals. In addition, the continuously developing practical strategies (e.g., activation functions~\cite{XuWCL15,nair2010rectified,he2015delving}, regularization~\cite{srivastava2014dropout,Sung:1996:LES:929901,IoffeS15}, and optimization~\cite{duchi2011adaptive,hinton2012neural,KingmaB14}) further contribute to the efficacy of deep neural networks.

Several region-based detectors depend on the strong classification capability of deep convolutional networks to evaluate generated RoIs. R-CNN is the first to adopt this approach by evaluating each RoI separately. Fast R-CNN~\cite{girshick2015fast} improved this method by allowing computation sharing through projecting RoIs to a shared feature map (called RoIPool layer, derived from SPPnets~\cite{he2014spatial}), resulting in better speed and accuracy. It was then integrated with the region proposal module (the Region Proposal Network, RPN) by sharing their convolutional features and extended to a unified network with "attention"~\cite{BahdanauCB14} mechanism, leading to further speedup and accuracy enhancement. R-FCN~\cite{li2016r} eliminates the fully-connected layers of region-based detectors and turns the whole model fully convolutional with the backbones of state-of-the-art image classifiers~\cite{he2016deep,szegedy2016rethinking} to fully share computation, contributing to a significant speedup. Mask R-CNN~\cite{he2017mask}, which adds a small Fully Convolutional Network (FCN)~\cite{long2015fully} as a parallel branch to standard Faster R-CNN and replaces the RoIPool layer with the RoIAlign layer, is the latest descendant of this stream and achieves significant advancement in several benchmarks of both the detection and segmentation tasks. However, most of these models use the multitask loss with equal weight settings without considering the influence of different loss type throughout the training process.

Recent work has focused on the post-recognition level of region-based detection models to boost the localization accuracy. Gidaris et al.~\cite{gidaris2015object} proposed a CNN-model for bounding box regression, which is used with iterative localization and bounding box voting. LocNet~\cite{gidaris2016locnet} aims to enhance the localization accuracy by assigning a probability to each border of a loosely localized search region for being related to the object's bounding box. It's different from the bounding box regression approaches~\cite{felzenszwalb2010object} adopted by most of the aforementioned region-based detectors and can be served as an effective alternative.

However, little work has focused on the advancement of region-based detectors from a data perspective. Online Hard Example Mining (OHEM)~\cite{shrivastava2016training} integrates bootstrapping~\cite{Sung:1996:LES:929901} (or {\em hard example mining}) with region-based detectors for a small extra computational cost, but still lacks enough focus on the localization accuracy because of the derived multi-task loss imbalance. Further discussion is available in Sect. 3.

\section{Model Design}
In this section, we argue that the current way of choosing hard examples lacks enough focus on localization accuracy and is suboptimal, and we will show that our approach results in better training (lower training loss), higher localization performance, and higher average precision. Firstly, we discuss the design motivation. Then we give a brief introduction of stratified sampling and definition of stratified constraint in this work. Finally, we present the design and implementation of our {\em Stratified Online Hard Example Mining} algorithm (S-OHEM).

\subsection{Motivation}
Most of the region-based detectors derive the multitask learning from Fast R-CNN, and assume equal contributions of classification loss and localization loss throughout the training process. However, this assumption is not often the case. We apply the original OHEM on standard Fast R-CNN and Faster R-CNN, then report the classification and localization loss throughout the training process on PASCAL VOC and KITTI datasets separately.

As is illustrated in Fig. 2, the classification loss is consistently larger than the localization loss (more than double in average). But this could result in a problem. Let's consider a situation where we have two region proposals RoI $A$ and RoI $B$ and are asked to choose one as the hard example for backpropagation. Based on the preliminary experiment result shown in Fig. 2, we make a common assumption that the training loss for RoI $A$ and RoI $B$ is $L_{cls}(A)=0.21$, $L_{loc}(A)=0.11$, and $L_{cls}(B)=0.19$, $L_{loc}(B)=0.12$ respectively. Recall that the classification loss is defined as log loss $L_{cls}(p,u)=-$log$p_u$ for true class $u$~\cite{girshick2015fast}, and thus the probability for the true class is 61.5\% and 64.5\% for RoI $A$ and RoI $B$ respectively. It's not a significant gap of the class prediction probability between these two RoIs, and we can believe they have similar performance for the classification task.

\begin{figure}
\begin{center}
\includegraphics[height=5cm]{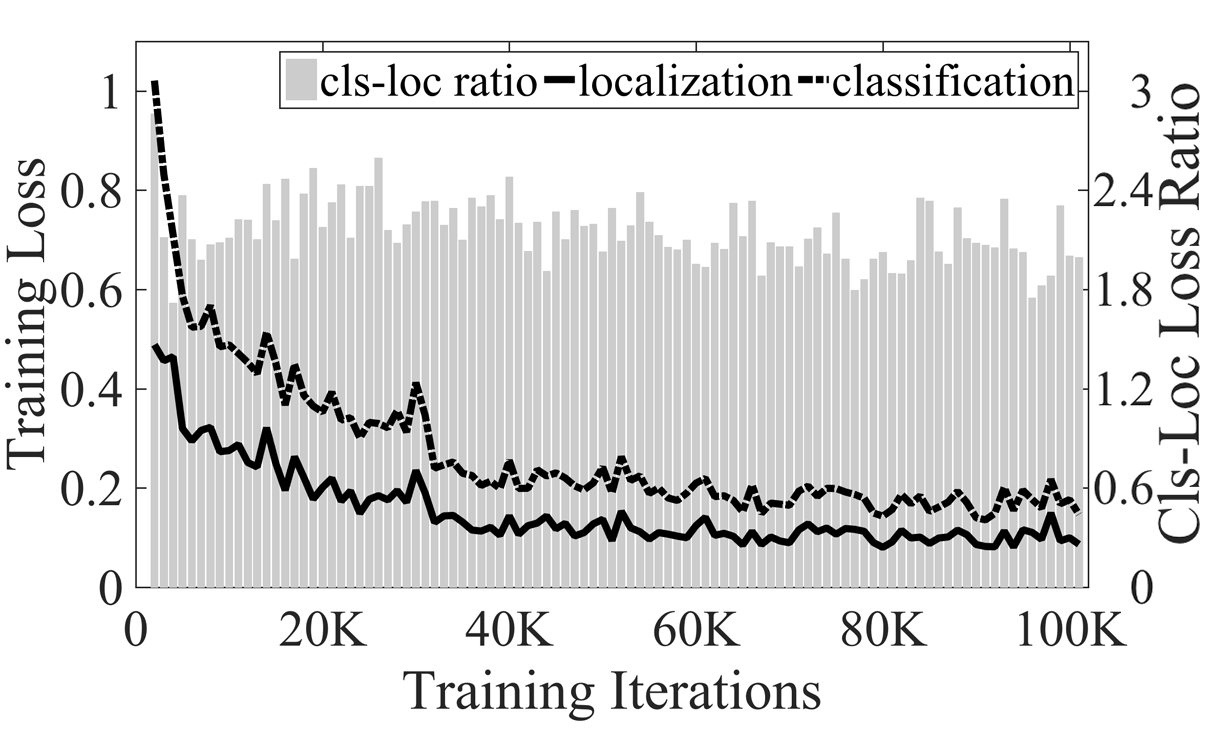}
\end{center}
   \caption{{\bf Influence of different loss types throughout the training process.}  
   For better visualization, we average out the training loss of every 1000 iterations.}
\label{fig:loss_iter}
\end{figure}

Regarding the localization loss, the gap between RoI $A$ and RoI $B$ is 0.01 $(L_{loc}(B) - L_{loc}(A) = 0.12 - 0.11 = 0.01)$. Within the smooth $L_{1}$ loss settings~\cite{girshick2015fast}, this gap turns to a 0.14 difference between the bounding boxes of ground truth and prediction. Note that this gap is quite significant when we use the parameterization for bounding box offsets given in~\cite{girshick2014rich}, and therefore we are supposed to choose RoI $B$ as the hard example for better localization accuracy and prediction quality. However, within the equal-weight multitask loss settings, RoI $A$ will be chosen as the hard one. Thus, the previous hard example mining approach lacks focus on localization accuracy.
\subsection{Stratified Sampling}
Stratified sampling is a sampling method involving the division of a population into distinct groups known as {\em strata}~\cite{li2016dss}. These strata are homogeneous subgroups of the original data with similar inner items. Stratified sampling can get higher statistical precision because the variability within subgroups sharing the same properties is lower than that of the entire population~\cite{thompson2012stratifiedsampling}. Therefore. stratified sampling improves the representativeness by reducing sampling error.

Each stratum constraint $s_k$ is denoted by $s_k = (p_k, f_k)$, where $p_k$ is a propositional formula and $f_k$ is the required sample size. In this work, the four stratum constraint is defined by the ratio between classification loss ($L_{cls}$) and localization loss: $s_1 = ($high $L_{cls}$ and high $L_{loc}, f_1)$, $s_2 = ($high $L_{cls}$ and low $L_{loc}, f_2)$, $s_3 = ($low $L_{cls}$ and high $L_{loc}, f_3)$, and $s_4 = ($low $L_{cls}$ and low $L_{loc}, f_4)$. The required sample size and threshold of high loss (hard examples) change dynamically throughout the training process.

\subsection{Stratified Online Hard Example Mining algorithm}
Given the observation that the previous hard example mining approach ignores the influence of different loss types throughout the training process and lacks focus on localization accuracy, we now demonstrate our approach of {\em Stratified Online Hard Example Mining} (S-OHEM).

The architecture of S-OHEM is shown in Fig 1. In each mini-batch iteration, S-OHEM firstly generates region proposals of the input images, forward-propagates all of them across the region-based detector, and gathers the training loss of each RoI. Then each RoI is assigned to one of the four strata defined in Sect. 3.2. Different loss type combinations represent how well the current detector performs in classification and localization tasks on each RoI respectively. Inside each stratum, hard examples are chosen by sorting the RoIs by loss. After that, all RoIs are subsampled according to a dynamic distribution, and a total number of $B$ hard examples are fed into the backpropagation process. The sampling distribution of RoIs from each stratum changes dynamically throughout the learning process, as each loss type maintains different contribution to the detector model at different training stages. Specifically, the effect of classification loss is more important in the beginning, while the localization loss contributes more at later training stages.

For implementation, we keep a record of history training loss and start to change the sampling distribution when the loss becomes stable (e.g., after 40K iterations shown in Fig 2). At the beginning of training, we only sample the first $B$ RoIs with high $L_{cls}$ (i.e., sample from $s_{12}$, the union of strata $s_1$ and $s_2$). When loss becomes stable, we gradually focus on choosing the RoIs with high $L_{loc}$ (i.e., sample from the union of $s_2$ and $s_3$, denoted by $s_{23}$) by increasing the sampling ratio between $s_{23}$ and $s_1$. Because of the gradually increasing focus on the localization loss, S-OHEM can predict more accurate bounding boxes and thus enhance the localization accuracy.

An equivalent alternative is available. To make it simple, we denote the contribution coefficient of $L_{cls}$ and $L_{loc}$ to hard example selection by $\alpha$ and $\beta$ respectively. And our approach aims to find the optimal value of $\alpha$ and $\beta$ in Formula (1) at different training stages. $L_{select}$ is only for hard example mining, and the actual loss backpropagated across the network will not be affected.

\begin{equation}
L_{select} = \alpha L_{cls} + \beta L_{loc}
\end{equation}

When training begins, we only sample the first $B$ RoIs with high $L_{cls}$ by setting $\alpha$ and $\beta$ in Formula (1) to 1 and 0 respectively. When loss becomes stable, we gradually focus on choosing the RoIs with high $L_{loc}$ by gradually decreasing the value of $\alpha$ and increasing $\beta$ in Formula (1). 

S-OHEM will not have a significant influence on the training time because most of the forward computation is shared between RoIs~\cite{girshick2015fast}, and the number of backpropagated examples is much smaller than that of all region proposals of the input images. To overcome co-located RoIs and loss double counting, we follow the solution of~\cite{shrivastava2016training} and apply non-maximum suppression (NMS)~\cite{neubeck2006efficient} to perform deduplication before the sampling procedure. NMS works by finding the highest loss RoI, and eliminating all other RoIs with lower loss and high overlap with the selected region. Besides, we derive their method of maintaining a read-only RoI network and a standard RoI network with sharing weights for efficient memory allocation. It is also worth noting that S-OHEM can be combined with any post-recognition regressors introduced in Sect. 2, because it focuses on enhancing the localization accuracy from the data perspective.

\section{Experiments and Results}
In this section, we conduct systematic experiments to evaluate the proposed S-OHEM and compare it with original OHEM. We describe the experimental setup in Sect. 4.1, and demonstrate the efficiency and accuracy of the algorithm by examining the training loss and average precision.

\subsection{Experimental Setup}
We use the standard and popular CNN architecture VGG16 from~\cite{SimonyanZ14a}, and evaluate the algorithms on the PASCAL VOC 2007 and KITTI Object Detection Evaluation 2012 dataset. In the PASCAL VOC experiment, training is done on the trainval set and testing on the test set. In the KITTI 2012 experiment, we use the first 5000 images to form the training set and the remaining 2481 images for testing. All models are trained with SGD for 80k mini-batch iterations and followed the same setup from Sect. 4.1. For average precision, we report the results with IoU thresholds of 0.5, 0.6, and 0.7, to evaluate the localization accuracy in a wider range of IoU thresholds. We use Fast R-CNN~\cite{girshick2015fast} as the detector base for our PASCAL VOC experiment, and Faster R-CNN~\cite{ren2015faster} for the KITTI 2012 experiment, to prove the usability of our approach. The initial learning rate is set to 0.001 and dropped in "steps" by a factor of 0.1 every 30K iterations. We process 2 images in each mini-batch iteration, and subsample 128 RoIs to feed them into backpropagation. Note that the baseline of OHEM reported in Table 2 (row 1-2) were reproduced and are slightly higher than the ones reported in~\cite{shrivastava2016training}.

For both experiments, we follow the procedure described in Sect. 3.3 to control the contribution coefficient of $L_{cls}$ and $L_{loc}$. In the beginning, $\alpha$ and $\beta$ are set to 1 and 0 when training starts. Then we gradually increase $\beta$ to the ratio between classification and localization loss when the loss becomes stable. Specifically, $\beta$ will increase to 1.9 and 2.3 for the VOC07 and KITTI12 experiment respectively.

\subsection{Results and Analysis}
\subsubsection{Training Convergence.}
We firstly analyze the training loss for both methods by logging the average training loss every 10K steps. Figure 3 shows the average loss per RoI for VGG16 with settings presented in Sect. 4.1. We see that S-OHEM yields lower loss in both classification and localization than the original OHEM, validating our claims that S-OHEM leads to better training than OHEM. Also, the results indicate that S-OHEM is better in classification confidence and localization accuracy during the training process.

\begin{figure}
\begin{center}
\includegraphics[height=5cm]{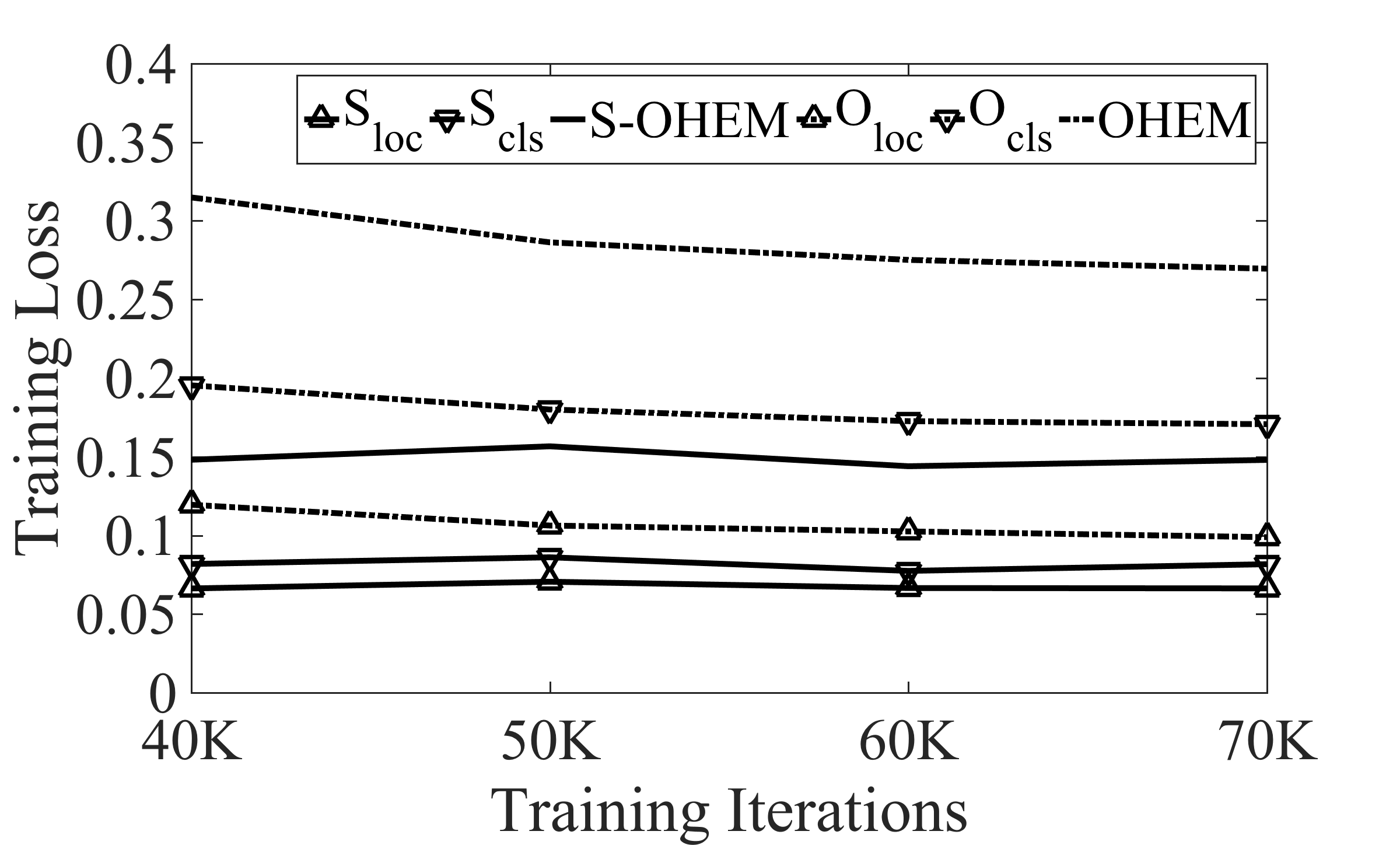}
\end{center}
\caption{{\bf Training loss for S-OHEM and OHEM.} We show the average loss per RoI for VGG16. These results indicate that S-OHEM is better in classification confidence and localization accuracy during the training process.}
\label{fig:training}
\end{figure}

\subsubsection{VOC 2007.}
Table 1 shows that on VOC07, S-OHEM improves the mAP of OHEM from 71\% to 71.1\% for an IoU threshold of 0.5, and an improvement of 0.4\% and 0.3\% for IoU 0.6 and 0.7 respectively. For category-specific improvements, S-OHEM performs well in most of the rigid categories (bold categories in Table 1) across all three IoU thresholds, especially for IoU 0.7.

As is listed on Table 3(a), we compute the mAP among rigid categories and show increase of 0.1\%, 0.5\%, and 0.5\% for IoU 0.5, 0.6, and 0.7 respectively. It's also interesting to find that S-OHEM performs quite well in detecting {\em cats} for IoU threshold 0.6, which indicates the better bounding boxes generated by S-OHEM in this environment.

\begin{table}
\begin{center}
\setlength{\tabcolsep}{0.2em} 
\resizebox{!}{0.57in}{%
\begin{tabular}{|c|c|c|cccccccccccccccccccc|}
\hline
method & IoU & mAP & {\bf aero} & {\bf bike} & bird & {\bf boat} & {\bf bottle} & {\bf bus} & {\bf car} & cat & {\bf chair} & cow & {\bf table} & dog & horse & {\bf mbike} & persn & plant & sheep & {\bf sofa} & {\bf train} & {\bf tv}\\
\hline\hline
OHEM 
& 0.5 & 71.0 & 72.1 & 80.4 & 68.9 & 60.5 & 47.1 & 81.5 & 79.6 & 82.8 & 54.1 & 77.3 & 70.7 & 81.7 & 81.4 & 76.7 & 74.4 & 41.6 & 70.0 & 69.6 & 76.5 & 73.6\\
S-OHEM 
& 0.5 & 71.1 & 72.8 & 80.9 & 69.2 & 60.2 & 47.9 & 81.4 & 79.5 & 82.5 & 53.8 & 76.6 & 70.3 & 81.9 & 81.5 & 77.5 & 74.5 & 41.6 & 70.1 & 70.2 & 76.0 & 73.6\\
improv 
& 0.5 & 0.1 & {\bf 0.7} & {\bf 0.5} & 0.3 & -0.3 & {\bf 0.8} & -0.1 & -0.1 & -0.3 & -0.3 & -0.7 & -0.4 & 0.2 & 0.1 & {\bf 0.8} & 0.1 & 0 & 0.1 & {\bf 0.6} & -0.5 & 0\\
\hline
OHEM
& 0.6 & 62.2 & 63.5 & 74.5 & 57.7 & 47.1 & 38.0 & 76.0 & 74.4 & 70.9 & 42.0 & 70.8 & 61.7 & 72.7 & 74.9 & 68.1 & 62.6 & 30.7 & 59.3 & 63.5 & 66.5 & 68.2\\
S-OHEM
& 0.6 & 62.7 & 64.9 & 74.4 & 58.5 & 48.1 & 38.5 & 76.6 & 73.9 & 75.3 & 42.0 & 71.8 & 60.3 & 72.8 & 74.7 & 69.2 & 62.0 & 30.8 & 59.1 & 65.0 & 67.6 & 68.5\\
improv
& 0.6 & 0.5 & {\bf 1.4} & -0.1 & {\bf 0.8} & {\bf 1.0} & {\bf 0.5} & {\bf 0.6} & -0.5 & {\bf 4.4} & 0 & {\bf 1.0} & -1.4 & 0.1 & -0.2 & {\bf 1.1} & -0.6 & 0.1 & -0.2 & {\bf 1.5} & {\bf 1.1} & 0.3
\\
\hline
OHEM
& 0.7 & 48.3 & 52.8 & 58.2 & 42.1 & 32.0 & 27.3 & 68.6 & 63.0 & 56.5 & 31.0 & 56.3 & 44.9 & 50.0 & 55.6 & 55.6 & 44.0 & 16.6 & 49.2 & 48.9 & 55.5 & 58.5\\
S-OHEM
& 0.7 & 48.6 & 55.2 & 57.8 & 41.4 & 32.5 & 27.9 & 69.0 & 63.8 & 56.9 & 30.4 & 58.2 & 44.9 & 50.0 & 54.1 & 56.1 & 44.2 & 16.4 & 48.6 & 49.9 & 56.2 & 58.9\\
improv
& 0.7 & 0.3 & {\bf 2.4} & -0.4 & -0.7 & {\bf 0.5} & {\bf 0.6} & {\bf 0.4} & {\bf 0.8} & {\bf 0.4} & -0.6 & {\bf 1.9} & 0 & 0 & -1.5 & {\bf 0.5} & 0.2 & -0.2 & -0.6 & {\bf 1.0} & {\bf 0.7} & {\bf 0.4}\\
\hline
\end{tabular}
}
\end{center}
\caption{{\bf VOC 2007 test detection average precision (\%).} All methods use VGG16 and bounding-box regression. Legend: {\bf IoU:} IoU threshold.}
\end{table}

\subsubsection{KITTI 2012.}
The evaluation results on KITTI 2012 is shown in Table 2. S-OHEM improves the mAP of OHEM from 63.9\% to 64.9\% for an IoU threshold of 0.6, and an improvement of 0.5\% for IoU 0.7. We also compute the mAP among rigid categories and list results in Table 3(b). Note that the Note that the {\em misc} category is classified as {\em rigid} based on our observation of the dataset. We show some increase of 1.6\% for both IoU thresholds 0.6 and 0.7. 

\begin{table}
\begin{center}
\resizebox{!}{0.6in}{%
\begin{tabular}{|c|c|c|ccccccc|}
\hline
method & IoU & mAP & {\bf car} & persn & cyclist & {\bf truck} & {\bf van} & {\bf tram} & {\bf misc}\\
\hline\hline
OHEM
& 0.5 & 78.5 & 78.5 & 62.9 & 72.4 & 87.5 & 89.9 & 87.6 & 71.0 \\
S-OHEM 
& 0.5 & 78.5 & 78.2 & 63.7 & 73.9 & 88.3 & 89.0 & 85.3 & 70.8 \\
improv 
& 0.5 & 0 & -0.3 & {\bf 0.8} & {\bf 1.5} & {\bf 0.8} & -0.9 & -2.3 & -0.2\\
\hline
OHEM
& 0.6 & 63.9 & 68.7 & 47.4 & 57.7 & 73.8 & 79.0 & 70.8 & 49.6 \\
S-OHEM
& 0.6 & 64.9 & 68.3 & 48.8 & 55.7 & 77.7 & 78.2 & 73.4 & 52.4 \\
improv
& 0.6 & 1 & -0.4 & {\bf 1.4} & -2 & {\bf 3.9} & -0.8 & {\bf 2.6} & {\bf 2.8}\\
\hline
OHEM
& 0.7 & 42.9 & 50.5 & 29.1 & 37.3 & 52.7 & 59.8 & 42.0 & 28.8\\
S-OHEM
& 0.7 & 43.4 & 49.8 & 28.8 & 33.3 & 61.2 & 60.1 & 39.3 & 31.7\\
improv
& 0.7 & 0.5 & -0.7 & -0.3 & -4 & {\bf 8.5} & 0.3 & -2.7 & {\bf 2.9}\\
\hline
\end{tabular}
}
\end{center}
\caption{{\bf KITTI 2012 test detection average precision (\%).} All methods use VGG16 and bounding-box regression. Legend: {\bf IoU:} IoU threshold.}
\end{table}

\begin{table}
\begin{center}
\setlength{\tabcolsep}{0.6em} 
\resizebox{!}{0.7in}{%
\begin{tabular}{ccc}
\begin{tabular}{|c|c|cc|}
\hline
method & IoU & rigid & non-rigid\\
\hline\hline
OHEM 
& 0.5 & 70.2 & 72.2\\
S-OHEM 
& 0.5 & 70.3 & 72.2\\
improv 
& 0.5 & 0.1 & 0\\
\hline
OHEM
& 0.6 & 62.0 & 62.4\\
S-OHEM
& 0.6 & 62.5 & 63.1\\
improv
& 0.6 & {\bf 0.5} & {\bf 0.7}\\
\hline
OHEM
& 0.7 & 49.7 & 46.2\\
S-OHEM
& 0.7 & 50.2 & 46.2\\
improv
& 0.7 & {\bf 0.5} & 0\\
\hline
\end{tabular}
&
\begin{tabular}{|c|c|cc|}
\hline
method & IoU & rigid & non-rigid\\
\hline\hline
OHEM 
& 0.5 & 82.9 & 67.7\\
S-OHEM 
& 0.5 & 82.3 & 68.8\\
improv 
& 0.5 & -0.6 & {\bf 1.1}\\
\hline
OHEM
& 0.6 & 68.4 & 52.6\\
S-OHEM
& 0.6 & 70.0 & 52.3\\
improv
& 0.6 & {\bf 1.6} & -0.3\\
\hline
OHEM
& 0.7 & 46.8 & 33.2\\
S-OHEM
& 0.7 & 48.4 & 31\\
improv
& 0.7 & {\bf 1.6} & -2.2\\
\hline
\end{tabular}\\
(a)&(b)
\end{tabular}
}
\end{center}
\caption{{\bf Category specific mean average precision (\%).} All methods use VGG16 and bounding-box regression. Legend: {\bf IoU:} IoU threshold. (a) On VOC 2007 test set. (b) On KITTI 2012 test set.}
\end{table}

\subsubsection{Rigid and Non-Rigid Category.}
Our experimental results have shown that S-OHEM performs quite well on rigid categories of both the VOC07 and KITTI12 dataset. The reason is that rigid bodies can reach better classification accuracy on pre-trained deep convolutional networks ascribed to its strong resistance to deformation. Therefore, the influence of different loss distribution throughout the training process (as described in Sect. 3.1) is more likely to happen on rigid bodies. Also, the border distribution of rigid bodies is more similar to each other and is thus easier to learn.

\section{Conclusion}

In this paper, we proposed Stratified Online Hard Example Mining (S-OHEM) algorithm, a simple and effective method for training region-based deep convolutional network detectors to enhance localization accuracy. During hard example mining, S-OHEM exploits stratified sampling and focuses on the influence of different loss types throughout the training process. Experimental analysis shows that S-OHEM outperforms OHEM regarding training convergence and localization accuracy, and achieves some AP improvements on {\em rigid categories} of PASCAL VOC 2007 and KITTI 2012. Besides, S-OHEM addresses the localization enhancing problem merely from the data perspective and can be easily plugged into existing region-based detectors. Furthermore, the state-of-the-art Mask R-CNN~\cite{he2017mask} also derives the equal-weight multi-task loss with an addition task of semantic segmentation, which is improvable through S-OHEM. S-OHEM can also be applied to other multi-task loss, including the loss of semantic segmentation, key-point detection, etc.

{\small
\bibliographystyle{ieee}
\bibliography{egbib}
}

\vfill
\end{document}